\documentclass[letterpaper]{article}
\usepackage{uai2019}
\usepackage[margin=1in]{geometry}
\pdfoutput=1

\usepackage{graphicx,color,amsmath,amsfonts,amssymb}
\usepackage{algorithm,algpseudocode}

\usepackage[backend=bibtex,citestyle=authoryear,bibstyle=authoryear]{biblatex}
\def\class{q}
\def\dpnts{\mathbf{X}}
\def\dpnt{\mathbf{x}}
\def\wghtsone{w}
\def\wghts{\mathbf{w}}
\def\momone{p}
\def\mom{\mathbf{p}}
\def\lbl{t}
\def\data{D}
\def\cost{J}
\def\mass{m}
\def\velonei{\frac{d\wghtsone_i}{dt}}
\def\vel{\mathbf{\frac{d\wghts}{dt}}}
\def\prob{\mathrm{prob}}

\def\model{M}

\def\trans{\intercal}
\def\kk{\mathbf{k}}
\def\kkone{k}
\def\ev{Z}

\title{Bayesian Neural Networks at Finite Temperature}

\author{ {\bf Robert J.N. Baldock, Nicola Marzari } \\
Theory and Simulation of Materials (THEOS) \\
and National Centre for Computational Design and Discovery of Novel Materials (MARVEL)\\
\'Ecole Polytechnique F\'ed\'erale de Lausanne, CH-1015 Lausanne, Switzerland\\
\texttt{rjnbaldock@gmail.com}
}

\date{April 2019}

\begin{document}
\maketitle

\begin{abstract}

We recapitulate the Bayesian formulation of neural network based classifiers and show that, while sampling from the posterior does indeed lead to better generalisation than is obtained by standard optimisation of the cost function, even better performance can in general be achieved by sampling finite temperature ($T$) distributions derived from the posterior. 
Taking the example of two different deep (3 hidden layers) classifiers for MNIST data, we find quite different $T$ values to be appropriate in each case.
In particular, for a typical neural network classifier a clear minimum of the test error is observed at $T>0$.
This suggests an early stopping criterion for full batch simulated annealing: cool until the average validation error starts to increase, then revert to the parameters with the lowest validation error. 
As $T$ is increased classifiers transition from accurate classifiers to classifiers that have higher training error than assigning equal probability to each class. 
Efficient studies of these temperature-induced effects are enabled using a replica-exchange Hamiltonian Monte Carlo simulation technique. 
Finally, we show how thermodynamic integration can be used to perform model selection for deep neural networks. 
Similar to the Laplace approximation, this approach assumes that the posterior is dominated by a single mode. 
Crucially, however, no assumption is made about the shape of that mode and it is not required to precisely compute and invert the Hessian.

\end{abstract}

\section{BAYESIAN FORMULATION OF NEURAL NETWORK CLASSIFIERS}

The Bayesian formulation of neural network classifiers was introduced by David MacKay~\parencite{mackay1992practical}.
In training a classifier one typically uses a minimisation algorithm to find a local minimum of a cost function defined over the parameters of the neural network model. 
If we use a softmax then the activation of the output units sum to one, and can be interpreted as the probability, assigned by the network, to the assertion that the input belongs to each individual class.
We can write this as $\prob\left(\class | \dpnt, \wghts \right)$ where $\class$ is the index of the class, $\dpnt$ is the single data example to be classified and $\wghts$ represents the parameters of the neural network.
For data in the training set $\data = \{\dpnt_i,\lbl_i\}$ where $\dpnt_i$ is an input data example and $\lbl_i$ the corresponding class label, then the probability that the network classifies $\dpnt_i$ correctly is $\prob\left(\class = \lbl_i | \dpnt_i, \wghts \right)$.
This probability is a function of the weights $\wghts$.
The probability that the network classifies the complete data set correctly is called the {\em likelihood} of the data, $\prob\left(\data|\wghts\right)$, and is given by $ \prod_i{\prob\left(\class = \lbl_i | \dpnt_i, \wghts \right)} $.

In general, the posterior probability for the weights, which expresses what we have learned about the weights from the data, is given by
\begin{align} 
\prob\left( \wghts | \data \right) =& \frac{ \prob\left(\data|\wghts\right) \prob\left(\wghts\right) }{ \prob\left(\data\right) } \label{eq:posterior_basic} \\
\propto&  \left[ \prod_i{\prob\left(\class = \lbl_i | \dpnt_i, \wghts \right)} \right] \prob\left(\wghts\right)  \label{eq:posterior_general}
\end{align}
where $\prob\left(\wghts\right)$ is the prior probability distribution for the weights, and represents our initial state of knowledge before the collection of any data.

If we assign a Gaussian prior with standard deviation $\sigma$ to the parameters $\wghts$ then the posterior~\eqref{eq:posterior_general} is given by
\begin{equation} \label{eq:posterior_gauss}
\prob\left( \wghts | \data \right) \propto \left[ \prod_i{\prob\left(\class = \lbl_i | \dpnt_i, \wghts \right)}\right]  e^{- \frac{\wghts^2}{2\sigma^2} }.
\end{equation}
The posterior probability of the weights~\eqref{eq:posterior_gauss} is therefore maximised when we minimise 
\begin{align} 
\cost(\wghts | \data) =& - \log{ \left[ \prob\left(\data|\wghts\right) \prob\left(\wghts\right)  \right]  } \label{eq:def_cost}  \\
 =& -\sum_i{\log\left[ \prob\left(\class = \lbl_i | \dpnt_i, \wghts \right)  \right]} + \frac{\wghts^2}{2\sigma^2}. \label{eq:intro_cost}
\end{align}
which can immediately be recognised as the cross entropy loss function with L2 regularisation.

In this paper we will use a uniform prior for $\wghts$
\begin{equation} \label{eq:uni_prior}
\prob\left(\wghts\right) = 
\begin{cases}
\frac{1}{\prod_{i=1}^d{\sigma_i}},  &|\wghtsone_i|<\sigma_i/2 \, \forall\: i, \\
0, &\mathrm{Elsewhere}.
\end{cases}
\end{equation}
Within the bounds of this prior the cost function is given by
\begin{equation} \label{eq:intro_cost_uniprior} 
\cost(\wghts | \data) = -\sum_i\left(\log\left[ \prob\left(\class = \lbl_i | \dpnt_i, \wghts \right)  \right] - \log{\sigma_i} \right).
\end{equation}
Outside the bounds of the prior the cost function is infinite. 
Minimising the cost function will simply mean maximising the likelihood within these bounds.

\section{ENERGY AND TEMPERATURE}

One may define the {\em potential energy} over the parameters of the network
\begin{align}
E(\data | \wghts) =& -\log{\prob\left( \data | \wghts \right)} \\
=& -\sum_i{\log\left[ \prob\left(\class = \lbl_i | \dpnt_i, \wghts \right)  \right]} \label{eq:def_e}
\end{align}
and the {\em ``temperature-adjusted''} posterior distribution
\begin{equation}
\prob\left( \wghts | T, \data \right) \propto e^{- \frac{1}{T}E(\data | \wghts) }\, \prob(\wghts)  . \label{eq:thermal_posterior}
\end{equation}
Here $T$ controls the noise in our posterior.
We name $T$ the {\em ``temperature''} in analogy with thermodynamic temperature.

For this paper, it is important to imagine how the temperature-adjusted posterior behaves as we vary $T$.
The distribution~\eqref{eq:thermal_posterior} tends towards the prior distribution for $T \rightarrow \infty$, and is equal to the true posterior distribution~\eqref{eq:posterior_general} for $T = 1$. 
At $T< 1$ (low temperature) the temperature-adjusted posterior~\eqref{eq:thermal_posterior} is concentrated around the maximum likelihood solution (the minimum of $E(\data| \wghts )$). 

\subsection{TEMPERATURE VS BATCH SIZE}

It is known that mini-batch learning improves generalisation.
Recently there has been much interest in how the batch size in mini-batch learning controls the noise level during learning, and under what conditions the optimiser can be said to be approximately sampling from the posterior~\parencite{smith2017bayesian, lengevin_md_NN, mandt2017stochastic, ahn2012bayesian,blundell2015weight}.
In this paper we instead use temperature to precisely control the noise in full batch training and investigate whether $T>0$ might be used as an alternative to early stopping for training neural network classifiers.

\section{ON THE OPTIMAL TEMPERATURE FOR BAYESIAN NEURAL NETWORKS}

Here we examine the average training and test errors of two deep neural network classifiers with parameters sampled from the temperature-adjusted posterior~\eqref{eq:thermal_posterior}, across a wide range of temperatures.
We find that sampling from the posterior ($T=1$) indeed leads to better test error than standard optimisation of the cost function.
However, we also find that in general, improved performance can be obtained at temperatures other than $T=1$ and that the optimal value of $T$ depends on the neural network and the data set.

\subsection{DATA AND NEURAL NETWORK ARCHITECTURES \label{sec:data_NN}}

\begin{figure}
\centering
  \includegraphics[width=0.5\linewidth]{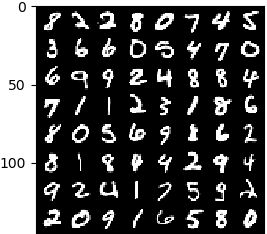}
\vspace{-1.0em}  \caption{ 64 random data samples from the MNIST training set, transformed as described in the text. 
}
  \label{fig:data_example}
\end{figure}

All calculations reported in this paper are performed using the MNIST data set~\parencite{lecun_mnist}.
For computational speed the MNIST data images are transformed down from 28 x 28 pixels to 16 x 16 (256 input dimensions).
Standard normalisation is applied to the full MNIST data set.
A random selection of rescaled data is shown in Figure~\ref{fig:data_example}.

Calculations are performed on stratified samples from the full MNIST training set.
Remaining samples from the MNIST training set, not included in these reduced stratified training sets, are appended to the MNIST test set. 
We adopt the convention that a data set containing $n$ data points is denoted $\data_n$.
Thus $\data_{500}$ is a data set of 500 images, 50 for each class.
Each time a data set of a new size is generated, it is stored.
Thus all calculations using data sets of the same size make use of exactly the same images.
These data sets are available from an open source repository, together with the code required to run these calculations~\parencite{nn_sample}.

In this section we make use of two neural networks.
The first, $\model^{(3)}$, has 256 input neurons, 3 hidden layers of 40 logistic neurons, and an output layer of 10 linear neurons to which a softmax is applied.
Thus, the loss function for this network is the cross-entropy loss function.
A second network $\model^{(3*)}$ has the same structure as $\model^{(3)}$, except that the final layer of linear neurons is replaced by a layer of {\em logistic} neurons.
A softmax is also applied to the output of these logistic neurons, so the loss function for this network is also the cross entropy loss function.
The use of logistic neurons in the final layer drastically reduces the capacity of $\model^{(3*)}$ to express probability distributions over the class labels as compared to $\model^{(3)}$.
In $\model^{(3*)}$ the values entering the softmax are constrained to the interval $(0,1)$.
Consequently, the smallest average training error that $\model^{(3*)}$ can achieve is $-\log\left( \frac{e^1}{e^1 + 9e^0} \right)\approx 1.46 $.
Conversely, the exact same network parameters $\wghts$ in $\model^{(3)}$ would yield an error of $E(\data|\wghts) = 0$.

\subsection{CHOICE OF PRIOR \label{sec:prior}}

The standard procedure for initialising the weights and biases of a neural network before training is to choose them uniformly at random inside the region $[-\frac{1}{\sqrt{k_i}},\frac{1}{\sqrt{k_i}}]$.
Here $i$ is an integer labelling the weight or bias, and $k_i$ is the fan in, including the bias, of the neuron to which weight or bias $i$ points.

In this paper we use a (uniform) prior~\eqref{eq:uni_prior} of this same shape, but made $50$ times wider for $\model^{(3)}$ and $1000$ times wider for $\model^{(3*)}$.
Including the factor $\frac{1}{2}$ from~\eqref{eq:uni_prior} for $\model^{(3)}$ we therefore have
\begin{equation} \label{eq:prior_used31}
\sigma_i = \frac{100}{\sqrt{k_i}}.
\end{equation}
Similarly, for $\model^{(3*)}$ the prior is as in~\eqref{eq:prior_used31}, but with a factor $2000$ in place of $100$.

These factors are chosen to create priors that capture important minima of $E(\data|\wghts)$, without being overly specific.
We choose the factors $50$ and $1000$ because, after standard initialisation followed by minimisation we found $\max_i{\left(\wghtsone_i\div\frac{1}{\sqrt{k_i}}\right)}$ to be 30 for $\model^{(3)}$ and 890 for $\model^{(3*)}$.
We repeated this procedure 25 times for each network to be sure these are typical values.
Minimisation is performed using Alg.~\ref{alg:rough_minimise} (see Sec.~\ref{sec:rehmc}), which is related to the FIRE optimisation algorithm~\parencite{FIRE}. 

\subsection{METHODS\label{sec:REMD}}

We develop a Hamiltonian (also called Hybrid) Monte Carlo (HMC)~\parencite{duane1987hybrid,neal2011mcmc}  formulation of a technique called replica-exchange molecular dynamics (REMD)~\parencite{REMD, REMD2} for use with neural network models.
This method enables efficient exploration of the temperature-adjusted posterior~\eqref{eq:thermal_posterior} across a range of temperatures, far faster than sampling those temperatures independently.

In replica-exchange HMC (RE-HMC) the user specifies a temperature range to study.
Normally this range goes from the lowest temperature of interest, up to an artificially high temperature where the sampler moves very quickly.
Independent samplers are created at particular temperatures throughout that range.
These samplers explore $\wghts$ space according to~\eqref{eq:thermal_posterior} at their respective temperatures.
Periodic attempts are made to exchange $\wghts$ vectors between neighbouring temperatures.
This accelerates $\wghts$ space sampling at low temperatures because each parameter vector spends some of the time at high temperature where it explores $\wghts$ space much more rapidly.
Importantly, samplers at each temperature are always sampling from the respective temperature-adjusted posterior~\eqref{eq:thermal_posterior} even immediately after a parameter swap.

A geometric series of temperatures is often employed for REMD in large parameter spaces.
This is known~\parencite{theor_geom_T, REMD2,empiric_geometric_T2} to yield nearly equal acceptance rates between neighbouring temperatures, unless the problem is pathological.

\subsubsection{Hamiltonian Monte Carlo\label{sec:hmc}}

HMC unifies two schools of statistical sampling: Markov chain Monte Carlo (MCMC)~\parencite{metropolis_MC} and Molecular Dynamics (MD)~\parencite{md_orig}.
MD introduces the concept of forward momentum $\mom$ and makes use of the gradient of the cost function.
Together $\mom$ and the gradient information guide $\wghts$ rapidly through complex $E(\data | \wghts )$ landscapes.
Standard MCMC makes no use of $\mom$ or the gradient: instead MCMC makes small random moves and diffuses inefficiently through $\wghts$ space.
However, if an intelligent proposal is known, perhaps a move between regions of state space with large weight in the temperature-adjusted posterior distribution, but separated by long distances and barriers of low probability density, then MCMC can be an extremely powerful approach.

HMC combines the benefits of MD and MCMC.
A short MD trajectory is performed to rapidly displace $\wghts$.
So long as the MD trajectory is reversible and symplectic, then 
this move can be treated as any other MCMC proposal.
Thus HMC enables the combination of intelligent MCMC moves with rapid MD trajectories.
HMC has been shown~\parencite{neal2011mcmc} to improve the scaling of sampling a $d-$dimensional multivariate Gaussian distribution from $d^2$ in MCMC to $d^{\frac{5}{4}}$.
The cost of HMC can further be reduced by a factor if one avoids backward motion between MD trajectories~\parencite{hmc_partial_mom,hmc_without_dbalance}. 
However, in Alg.~\ref{alg:HMC} we present the simplest HMC algorithm since this already improves the scaling, and $\model^{(3)}$ and $\model^{(3*)}$ have 10690 dimensions each.
Alg.~\ref{alg:HMC} assumes an uniform prior as in~\eqref{eq:intro_cost_uniprior}.
We use the Velocity Verlet scheme~\parencite{velocity_verlet_orig} to propagate the MD trajectories.

The forward momentum $\mom$ is related to the velocity $\vel$ by $\momone_i =\mass_i\velonei $ where $\mass_i$ is called the ``mass'' of $\wghtsone_i$.
In this paper $\mass_i=1\;\forall i$.
In general these $\mass_i$ can be set to different values, which may make sampling more efficient.
For example, one might set $\mass_i = k_i$, where $k_i$ is the fan in of the target neuron for $\wghtsone_i$ as described in Sec.~\ref{sec:prior}.
This would have the effect of approximately equalising the expected change in the input to each neuron for each time step, subject to the assumption that the activations of neurons in the preceding layer and the weights leading to each neuron are all independent. 

\begin{algorithm}[h]
\caption{One Hamiltonian Monte Carlo trajectory. \label{alg:HMC} Samples $\prob(\wghts) \propto e^{-\cost(\wghts|\data)}$ for a uniform prior as in~\eqref{eq:intro_cost_uniprior}.}
\begin{algorithmic}
\Procedure{HMC}{$\wghts$}
\State $\wghts_o = \wghts$ \Comment{Copy initial $\wghts$}
\State Draw random $\momone_{0,i} \sim \mathcal{N}(0,\mass_i T) \forall i$
\\ \hspace{3.0em} \Comment Random $\mom_o$
\State $U_o = E\left(\data|\wghts_o\right) + \sum_i{\frac{\momone_{o,i}^2}{2\mass_i}}$ 
\\ \hspace{2.0em} \Comment{$U_0$ is initial total energy}
\State Propagate $(\wghts_o , \mom_o)$ using the Velocity Verlet algorithm for $L$ time steps of length $dt$. Final coordinates are $(\wghts_n , \mom_n)$.
\State $U_n = E\left(\data|\wghts_n\right) + \sum_i{\frac{\momone_{n,i}^2}{2\mass_i}} $ \Comment{Final total energy}
\State $\alpha =  \frac{1}{T} \left( U_o - U_n \right) $ \Comment log acceptance probability
\State Draw $x \sim \mathcal{U}(0,1)$ \Comment Random number
\If{$(|\wghtsone_i|<\sigma_i/2 \,\forall\; i )$ AND $ (\log(x)<\alpha) $ }
\State $\wghts = \wghts_n$ 
\EndIf \\
\Return $\wghts$ \Comment Returns either initial $\wghts$ or $\wghts_n$
\EndProcedure

\end{algorithmic}
\end{algorithm}

\subsubsection{Replica-Exchange Hamiltonian Monte Carlo\label{sec:rehmc}}

Our RE-HMC algorithm proceeds as follows.
We used a geometric series of $N_T$ temperatures, as described in Sec.~\ref{sec:REMD}.
The current parameters values for the sampler at temperature $T_i$ are $\wghts_i$.
To save space, we will write $E_i = E\left(\data | \wghts_i \right)$.

\begin{algorithm}[t]
\caption{Replica-exchange Hamiltonian Monte Carlo \label{alg:REHMC}}
\begin{algorithmic}
\Procedure{RE-HMC}{$\{(T_i,\wghts_i)\}$}
\Loop
  \For{$T_i$ in $\{T\}$ }
    \State Update sample $\wghts_i$ from~\eqref{eq:thermal_posterior} by performing $N_\mathrm{traj}$ HMC trajectories of length $L$ and time step $dt$.
  \EndFor
  \For{$i \gets 1, N_T$}
    \State Choose random adjacent $T$ pair $\left(T_j, T_{j+1} \right)$.
    \State Exchange $\left( \wghts_j , \wghts_{j+1} \right)$ with probability
    \State $\min{\left[1,e^{\left( \frac{1}{T_j} - \frac{1}{T_{j+1}} \right) \left( E_j - E_{j+1} \right) }\right]}$.
  \EndFor
  
\EndLoop
\EndProcedure
\end{algorithmic}
\end{algorithm}

In this fresh implementation independent time steps $dt$ are set for each temperature by running additional HMC trajectories which are not included in the main simulation, and measuring the acceptance rate of the trajectories.
The time step is updated to obtain an acceptance rate inside a the range (0.6--0.7), which is centred around the optimal value for HMC sampling of a multivariate Gaussian distribution, $0.65$~\parencite{neal2011mcmc}.

At the start of the simulation the parameters of the neural network are initialised independently for each temperature using the following approach:
\begin{enumerate}
\item Draw $\wghts$ uniformly at random from the region $\wghtsone_i \in [-\frac{1}{\sqrt{k_i}},\frac{1}{\sqrt{k_i}}]$ with $k_i$ as defined in Sec.~\ref{sec:prior}.
\item Minimise $E(\data|\wghts)$ using Alg.~\ref{alg:rough_minimise}. 
This minimisation avoids starting the dynamics from parameter values where the gradient is extremely large.
\item Perform a number of burn-in trajectories at the appropriate temperature.
\end{enumerate}

\begin{algorithm}[t]
\caption{Fast minimisation of $E(\data|\wghts)$.\label{alg:rough_minimise}}
\begin{algorithmic}
\Procedure{RMin}{$\wghts$}

\State $\mom = 0$ \Comment{Set initial momenta to zero.}
\For{$i = 1, N_{\mathrm{steps}} $}
\State $\wghts_\mathrm{save},\, E_\mathrm{save} = \wghts,\, E(\data | \wghts)$ \Comment{Save lowest energy state.}
\State Propagate $(\wghts,\mom)$ through 1 Velocity Verlet time step, duration $dt$.
\If{$E(\data|\wghts)<E_\mathrm{save}$}\Comment{Downwards step}
\State $dt = dt + 0.05$.\Comment{Slowly increase $dt$.}
\Else\Comment{Upwards step}
\State $\mom = 0$ \Comment{Zero momenta}
\State $\wghts = \wghts_\mathrm{save}$\Comment{Return to lowest energy $\wghts$}
\State $dt = dt*0.95$ \Comment{Rapidly decrease $dt$.}
\EndIf
\EndFor
\EndProcedure
\end{algorithmic}
\end{algorithm}
Alg.~\ref{alg:rough_minimise} is related to the FIRE optimisation algorithm~\parencite{FIRE}.
We found Alg.~\ref{alg:rough_minimise} to be extremely efficient for these simple networks.
For $\data_{50}$ and $\data_{500}$ Alg.~\ref{alg:rough_minimise} required just a few hundred steps to reach $E(\data|\wghts)=0$ in $\model^{(3)}$.

\begin{figure}
  \includegraphics[width=\linewidth]{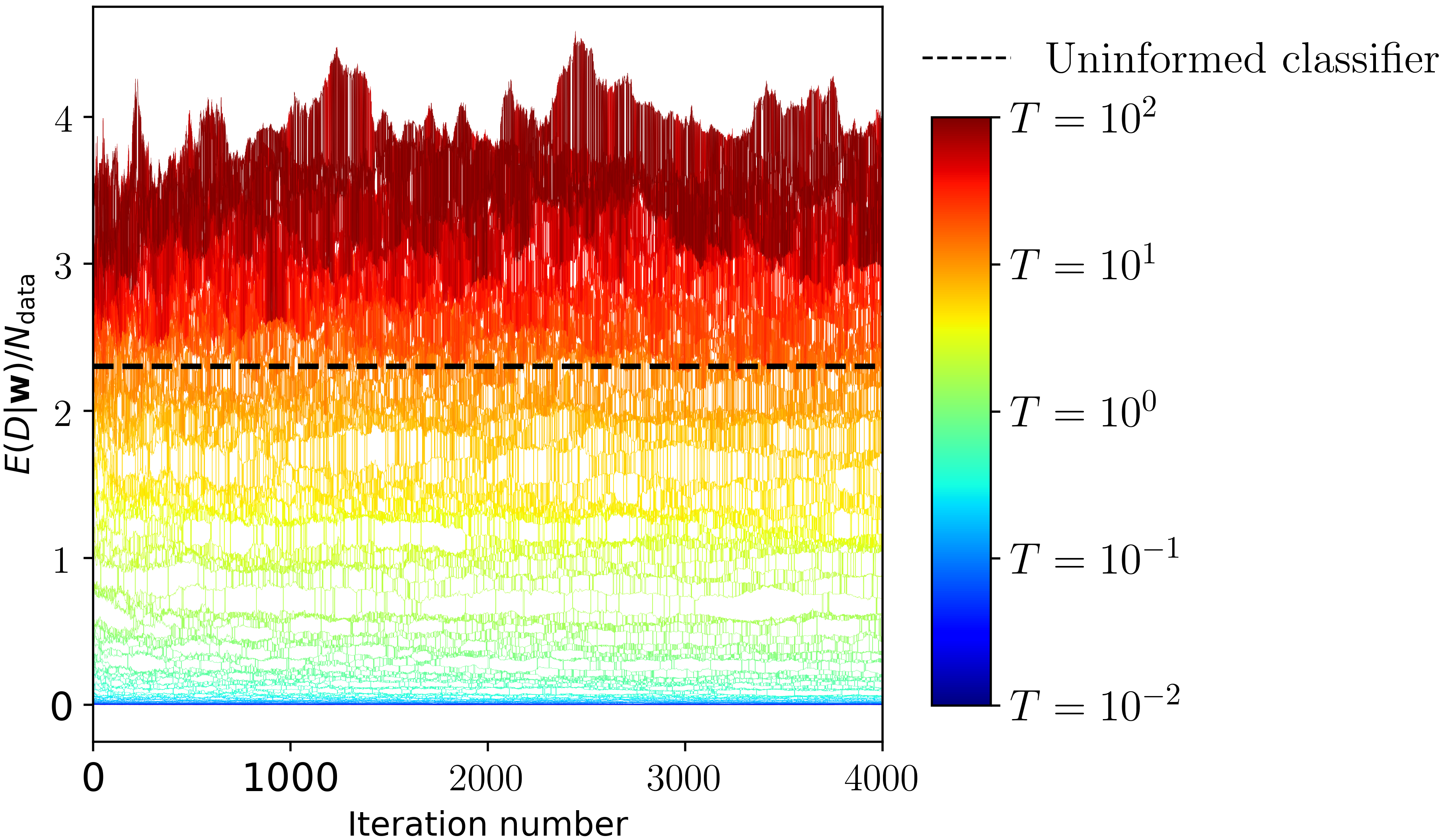}
  \caption{Trajectory of a replica-exchange HMC Simulation for $\model^{(3)}$ with $\data_{500}$. 
  \label{fig:pt_traj}
}
\end{figure}

Fig.~\ref{fig:pt_traj} shows the trajectory of a RE-HMC simulation with $\model^{(3)}$ and $\data_{500}$.
The parameters used are given in Table~\ref{table:re_hmc_params}.
The typical value of $E$ explores a narrow range which is specified by the temperature. 
At high temperature $\model^{(3)}$ has higher training loss than the test loss of an uninformed classifier that assigns equal probability to each digit class for every image.
From~\eqref{eq:def_e} it is apparent that such high training losses can be achieved by assigning low probability to the correct class label for just a fraction of training examples.

\begin{table}
\caption{RE-HMC Parameters in Figs.~\ref{fig:pt_traj},~\ref{fig:posterior_varT} and~\ref{fig:posterior_varT_M3star}.}
\label{table:re_hmc_params}
\begin{center}
\begin{tabular}{ccc}
\multicolumn{1}{c}{\bf PARAMETER}  &\multicolumn{1}{c}{\bf FIGS.~\ref{fig:pt_traj}  AND~\ref{fig:posterior_varT} } &\multicolumn{1}{c}{\bf FIG.~\ref{fig:posterior_varT_M3star}} \\
\hline \\
$N_T$         &112  & 84 \\
$T_\mathrm{min}$   &$10^{-2} $ & $10^{-\frac{3}{2}}$ \\
$T_\mathrm{max}$   &$10^{2} $ & $10^{\frac{3}{2}}$  \\
$N_\mathrm{traj}$  &10 & 10  \\
$L$           &100 & 100 \\
\end{tabular}
\end{center}
\end{table}

\subsubsection{Relationship to Previous Work \label{sec:rehmc_rel_prev_work}}

HMC was first introduced to Bayesian learning for neural networks by Radford Neal~\parencite{neal_thesis}.
Researchers have also applied a MD algorithm called Langevin dynamics~\parencite{lengevin_md_NN} to sample the posterior for neural networks. 
In contrast to HMC, samples drawn using Langevin dynamics are only drawn from the true posterior asymptotically in the limit of vanishing $dt$.
In~\parencite{chandra2018langevin} the authors use both replica-exchange with MCMC (also called ``parallel tempering'') and REMD with Langevin dynamics to accelerate exploration of neural network parameter spaces.
However, they do not investigate temperature effects. 
Instead, after performing replica-exchange simulations, they set $T=1$ for all samplers, irrespective of their temperatures during replica-exchange, and draw further samples from the posterior.
Parallel tempering has also been successfully applied to the training of restricted Boltzmann machines (e.g.~\parencite{cho2011improved, cho2010parallel, desjardins2010adaptive, desjardins2014deep}.)

To our knowledge, this is the first time a RE-HMC approach has been applied to study neural networks.
Different RE-HMC algorithms adapted to other scientific domains can be found in~\parencite{RE_HMC_MANIFOLD, RE_HMC_CHEM}.

\subsection{RESULTS: THE BEST GENERALISATION IS NOT ALWAYS FOUND AT $T=1$ \label{sec:post_small_data}}

There is broad agreement in the Bayesian learning community that sampling from the posterior leads to better generalisation than minimising the cost function.
Our results strongly support this assertion.
However, we also observe that improved performance can in general be obtained at temperatures other than $T=1$, and that the optimal value of $T$ depends on the network and the data set.

\begin{figure}
  \includegraphics[width=\linewidth]{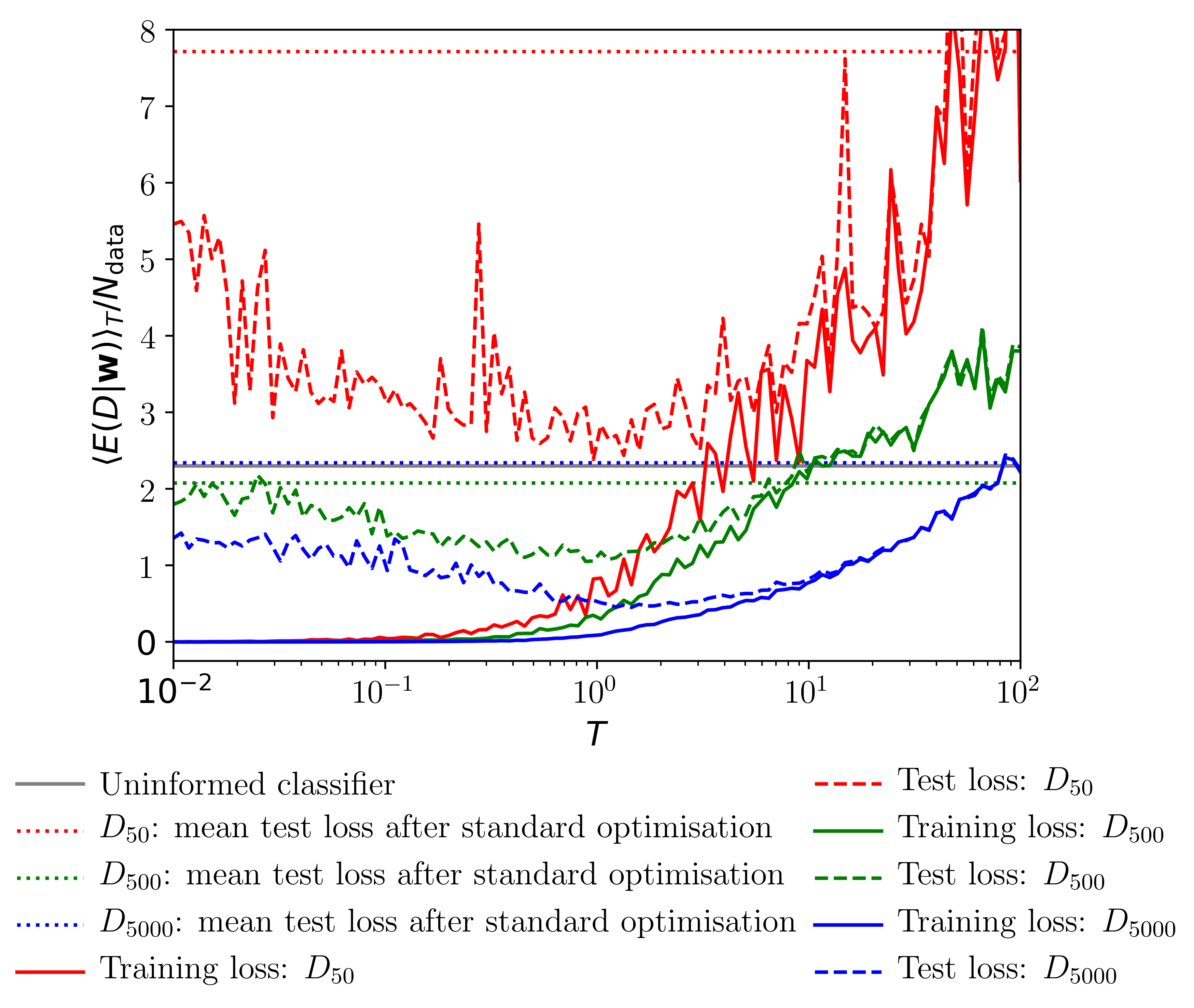}
  \caption{Empirical $\langle E(\data|\wghts) \rangle_{T}$ for $\model^{(3)}$ at different temperatures $T$. Sharp variations are due to sampling noise. ``Standard optimisation'' of $\model^{(3)}$ is described in the text.
  \label{fig:posterior_varT}
}
\end{figure}

Fig.~\ref{fig:posterior_varT} shows the sampled average of the potential energy $\langle E(\data|\wghts) \rangle_{T}$ for $\model^{(3)}$ across a wide range of $T$, and using different sized training sets.
The parameters of this calculation are shown in Table~\ref{table:re_hmc_params}.
In all data sets sampling from the posterior yields much improved generalisation as compared to standard optimisation. 
Here we apply the conventional initialisation described in Sec.~\ref{sec:prior} and minimise $E(\data|\wghts)$ using Alg.~\ref{alg:rough_minimise}.
For $D_{50}$ and $D_{500}$ we repeat the procedure until we obtain 100 vectors $\{\wghts\}$ with $E(\data|\wghts)=0$ for each data set (200 in total). 
Fig.~\ref{fig:posterior_varT} reports the mean test errors of these parameter sets. 
It is far more difficult to obtain $E(\data_{5000}|\wghts)=0$ by this optimisation method, so instead we repeated this optimisation 4000 times, keeping the 100 $\wghts$ with the lowest training error.
The reported test error for $D_{5000}$ is the mean test error of those 100 $\wghts$.

At low temperatures $\model^{(3)}$ achieves near zero training error on all data sets.
Conversely, clear minima are observed in the test error.
For the smaller data sets, $\data_{50}$ and $\data_{500}$, the minimum test error does indeed occur in the region of $T \sim 1$.
However, for the larger data set $\data_{5000}$ (which is still relatively small) that minimum occurs at $T>1$.

The $T$ of minimum test error corresponds to the $T$ value at which $\model^{(3)}$ begins overfitting the data.
At $T\gg1$ the network is essentially untrained.
Between the extremes of $T\ll1$ and $T\gg1$ the network learns, but not so much as to overfit the data.
This leads to the observed minimum.
Our results suggest a stopping criterion for full batch simulated annealing~\parencite{kirkpatrick1983optimization} of classifiers: {\em cool the network until the average validation error starts to increase, then revert to the parameters with the smallest validation error}.

At high $T$ and for all data sets, $\model^{(3)}$ has higher test loss than both $\wghts$ obtained by standard optimisation and an ``uninformed'' classifier, which assigns an equal probability of $0.1$ to each digit class, independent of the image.
As the data set is increased, the temperature at which the network test loss exceeds that of the ``uninformed'' classifier becomes higher. 
For the smallest data set $\data_{50}$ the network is in this state for all temperatures.

Finally we remark that in $\model^{(3)}$, a gradual transition from accurate to inaccurate classification of the training data occurs as $T$ is increased.
No sharp jump is observed: a statistical physicist would say that there is no ``first-order phase transition''.
If a sharp transition between accurate and inaccurate classifiers were observed then 
accurate and efficient
sampling of the temperature-adjusted posterior would 
require 
specialised techniques such as Metadynamics~\parencite{metadynamics} or an adaptive temperature replica-exchange algorithm~\parencite{feedback_opt_pt}.
From this perspective, the absence of a first-order phase transition is an important result, since it makes sampling the temperature-adjusted posterior distribution relatively simple and inexpensive.

\begin{figure}
  \includegraphics[width=\linewidth]{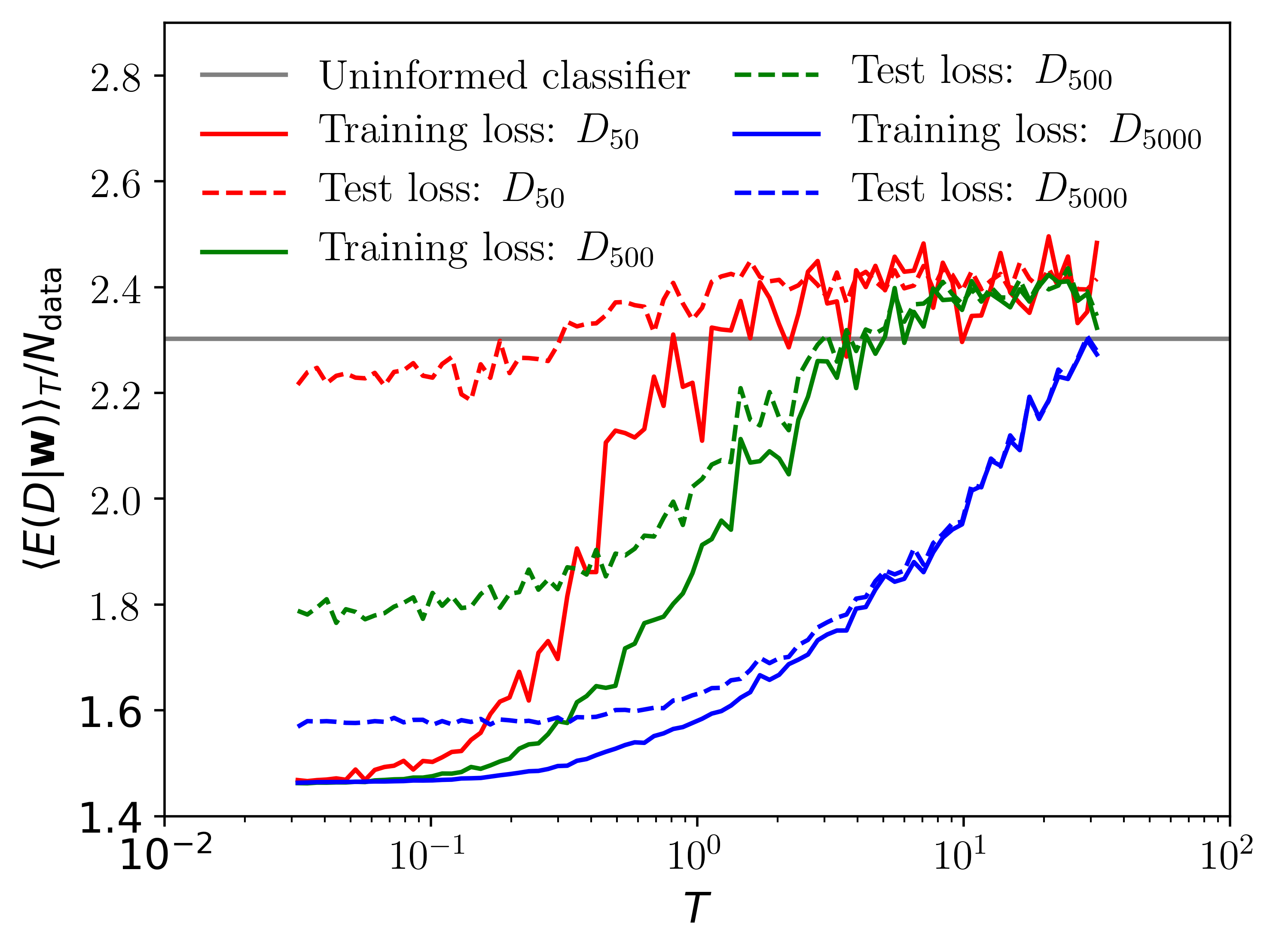}
  \caption{Empirical $\langle E(\data|\wghts) \rangle_{T}$ for $\model^{(3*)}$ across a range of temperatures $T$.
  Sharp variations are due to sampling noise.
  \label{fig:posterior_varT_M3star}
}
\end{figure}

In order to check the generality of these observations we repeated the calculations shown in Fig.~\ref{fig:posterior_varT} for the neural network $\model^{(3*)}$.
The results for $\model^{(3*)}$ are shown in Fig.~\ref{fig:posterior_varT_M3star}.
The parameters for this calculation can be found in Table~\ref{table:re_hmc_params}.

It can be seen from Fig.~\ref{fig:posterior_varT_M3star} that $\model^{(3*)}$ has lowest test error at $T\ll 1$: no minimum is apparent across the temperatures studied.
This is a manifestation of $\model^{(3*)}$'s low capacity for expressing the likelihood function.
As in $\model^{(3)}$, a transition is observed for $\model^{(3*)}$ from relatively accurate to inaccurate classification as $T$ is increased.
Overall $\model^{(3*)}$ generalises approximately as well as $\model^{(3)}$ for the smallest data set $\data_{50}$.
However, even for $\data_{500}$ the best test loss of $\model^{(3)}$ is already less than 1.46: the lowest value $\model^{(3*)}$ can achieve (see Sec.~\ref{sec:data_NN}).
We anticipate that this trend will continue.

\subsection{SUMMARY \label{sec:pt_conclusion}}

We find that, for $\model^{(3)}$, which is a typical neural network classifier, sampling the posterior distribution does indeed give lower test error than standard optimisation of the cost function.
However, even better performance can often be obtained at temperatures other than $T=1$.
Furthermore, quite different values of $T$ are appropriate to different neural networks and data sets.
The optimal temperature can be located by sampling the temperature-adjusted posterior across a wide range of temperatures to identify that with the lowest validation error.
For typical neural networks a clear minimum of the test error is observed at $T>0$.
This suggests a stopping criterion for full batch simulated annealing of classifiers: cool until the average validation error starts to increase, then revert to the parameters with the smallest validation error.
As temperature is increased we observe a gradual transition from accurate to inaccurate classifiers of the training data.
The absence of a sharp transition is significant for the choice of sampling method used.
It implies that no special temperature schedule is required for sampling methods such as the RE-HMC scheme presented here, and that more specialised methods such as Metadynamics, which are designed to enhance mixing across such sharp transitions, are not required.

\section{BAYESIAN MODEL SELECTION FOR DEEP (AND SHALLOW) NEURAL NETWORKS WITH THERMODYNAMIC INTEGRATION \label{sec:hess_free_mod_selec}}

In this section we demonstrate how thermodynamic integration (TI), can be used to perform Bayesian model selection between two neural network models: one deep and one shallow network.
We begin with a quick overview of Bayesian model selection before describing TI, the details of our calculation, and presenting the results. 

TI is related to annealed importance sampling~\parencite{neal_ais}.
Encouragingly, annealed importance sampling has successfully been applied to Boltzmann machines~\parencite{ais_boltzmann_machines}, deep belief nets~\parencite{ais_deep_belief_nets} and deep generative models~\parencite{ais_deep_generative_models}.

\subsection{NEURAL NETWORK ARCHITECTURES AND PRIORS \label{sec:ti_nn_and_prior}}

In this section we reuse the neural network $\model^{(3)}$, described in Sec.~\ref{sec:data_NN}.
We shall compare $\model^{(3)}$ to a second neural network, $\model^{(1)}$, which is identical to $\model^{(3)}$ except that, in place of the 3 hidden layers in $\model^{(3)}$, $\model^{(1)}$ contains only a single hidden layer.

The priors of these two networks are different because $\model^{(1)}$ has fewer parameters than $\model^{(3)}$.
Standard training of $\wghts$ for $\model^{(1)}$ using Alg.~\ref{alg:rough_minimise} gives $\max_i{\left(\wghtsone_i\div\frac{1}{\sqrt{k_i}}\right)} =33$ at $E(\data_{500}|\wghts) = 0$.
This is similar to the value 30 obtained for $\model^{(3)}$.
Consequently we used the same width factor 50 to define the prior for $\model^{(1)}$, as in $\model^{(3)}$, thereby assigning the bounds of the prior $\mathbf{\sigma}$ according to~\eqref{eq:prior_used31}.

\subsection{INTRODUCTION TO BAYESIAN MODEL SELECTION\label{sec:intro_bayes_mod_selec}}

Bayesian model selection for neural networks was first formulated by David MacKay~\parencite{mackay1992evidence, MacKay91bayesianinterpolation}.
Imagine we have two different machine learning models $\model_1$ and $\model_2$, and a data set $\data = \{\dpnt,\lbl_i\}$, or in vector notation $\data = \left(\mathbf{\dpnts}, \mathbf{\lbl} \right)$ where $\mathbf{\dpnts}$ is a matrix of all the input data and $\mathbf{\lbl}$ a vector of the corresponding class labels. 

In Bayesian model selection one computes and compares the probabilities of the models given the data, $ \prob\left(\model|\mathbf{\dpnts}, \mathbf{\lbl}\right)$.
A true Bayesian would believe both models to different extents, according to their probabilities.
However it is common practice to choose the model with highest probability, discarding all others.

The ratio of the probabilities for $\model_1$ and $\model_2$ can be found as follows
\begin{align}
\prob\left(\model, \mathbf{\lbl}|\mathbf{\dpnts}\right) =& \prob\left(\model|\mathbf{\dpnts}, \mathbf{\lbl}\right) \prob\left( \mathbf{\lbl}|\mathbf{\dpnts}   \right) \\
=& \prob\left(  \mathbf{\lbl}  | \mathbf{\dpnts}, \model \right) \prob\left( \model |\mathbf{\dpnts}   \right) \\
\Rightarrow\frac{\prob\left(\model_1|\mathbf{\dpnts}, \mathbf{\lbl}\right)}{ \prob\left(\model_2|\mathbf{\dpnts}, \mathbf{\lbl}\right)  } =& 
\frac{\prob\left(  \mathbf{\lbl}  | \mathbf{\dpnts}, \model_1 \right)}{\prob\left(  \mathbf{\lbl}  | \mathbf{\dpnts}, \model_2 \right)  }\frac{ \prob\left( \model_1 \right)}{\prob\left( \model_2 \right)} 
\label{eq:prob_M_ratio}
\end{align}

The term furthest to the right in~\eqref{eq:prob_M_ratio} is the ratio of our priors for the models and represents our initial bias towards either model.
Typically we have no initial preference between $\model_1$ and $\model_2$, in which case this term is equal to $1$.

In theory one can calculate $\prob\left(  \mathbf{\lbl}  | \mathbf{\dpnts}, \model \right)$ as follows
\begin{align}
\prob\left(  \mathbf{\lbl}  | \mathbf{\dpnts}, \model \right) =& \int{d\wghts \prob\left(  \mathbf{\lbl} , \wghts | \mathbf{\dpnts}, \model \right) } \\
 =& \int{d\wghts \prob\left(  \mathbf{\lbl}  |\wghts, \mathbf{\dpnts}, \model \right) \prob\left( \wghts \right)} \label{eq:int_posterior} \\
 =& \int{d\wghts e^{-\cost\left(\wghts|\data\right)}} \label{eq:integrate_cost}
\end{align}
In~\eqref{eq:integrate_cost} we have made a substitution from~\eqref{eq:def_cost}.

\subsection{METHOD: THERMODYNAMIC INTEGRATION\label{sec:ti}}

In this section we describe how~\eqref{eq:integrate_cost} can be calculated using TI.
The TI algorithm we describe here was originally introduced in~\parencite{frenkel_ti_crystal}.
This approach asserts that the integrand of~\eqref{eq:integrate_cost} is approximately unimodal, and localised in the region surrounding a particular local minimum of $\cost\left(\wghts|\data\right)$, at $\wghts_0$.

We then sample $e^{-\cost\left(\wghts|\data\right)}$ in the region of $\wghts_0$ and fit an approximate quadratic form for $\cost\left(\wghts|\data\right)$  
\begin{equation} \label{eq:quadrat_approx}
\frac{1}{2}\left(\wghts-\wghts_0\right)^\trans\kk \left(\wghts-\wghts_0\right) \simeq \cost\left(\wghts|\data\right) .
\end{equation}
The evidence for that quadratic form with a uniform prior
\begin{equation} \label{eq:def_Z0}
\ev_0 = \int_{|\wghts|<\frac{\mathbf{\sigma}}{2}}{e^{-\left(\wghts-\wghts_0\right)^\trans\frac{\kk}{2} \left(\wghts-\wghts_0\right)}} 
\end{equation}
is known exactly, provided $\kk$ is diagonal.
For this reason we fit a diagonal matrix $\kk$ with 
\begin{equation}\label{eq:fit_k}
\kkone_{ii} = \frac{1}{\langle (\wghtsone_i - \wghtsone_{0,i})^2 \rangle}.
\end{equation}
The average $\langle (\wghtsone_i - \wghtsone_{0,i})^2 \rangle$ is taken by sampling from the posterior distribution~\eqref{eq:posterior_general}.
For a uniform prior then it is important that, for this averaging alone, we should instead sample from $\prob(\wghts)\propto e^{-E(\data|\wghts)}$ without applying hard boundaries to $\wghts$.
This ensures that the approximation~\eqref{eq:quadrat_approx} is a good fit to $\cost\left(\wghts|\data\right)$.

Next in TI one sequentially samples a series of $N_\mathrm{bridge}$ bridging potentials constructed to interpolate between $\cost\left(\wghts|\data\right)$ and the approximation~\eqref{eq:quadrat_approx}.
\begin{align}\label{eq:bridge_cost}
& \cost_\mathrm{bridge}(\wghts | \data, \lambda) = (1-\lambda)( \cost(\wghts | \data)  - \cost(\wghts_0 | \data) ) \nonumber \\ 
&+ \lambda\left(\wghts-\wghts_0\right)^\trans\frac{\kk}{2}\left(\wghts-\wghts_0\right) + \cost(\wghts_0 | \data) .
\end{align}
Thus $\cost_\mathrm{bridge}(\wghts | \data, \lambda=0) = \cost(\wghts | \data)$ and $\cost_\mathrm{bridge}(\wghts | \data, \lambda=1) = \left(\wghts-\wghts_0\right)^\trans\frac{\kk}{2}\left(\wghts-\wghts_0\right)$.

Defining
\begin{align}
F =& - \log{ \int_{|\wghts|<\frac{\mathbf{\sigma}}{2}}{e^{-\cost(\wghts | \data) }} } \label{eq:def_FE} \\
F_0 =& - \log{ \ev_0 } \label{eq:def_FE0}
\end{align}
it can be shown~\parencite{frenkel2001understanding} that
\begin{equation} \label{eq:ti_integrate}
 F = F_0 + \int_{\lambda=1}^{\lambda=0}{d\lambda \biggl< \frac{\partial \cost_\mathrm{bridge}(\wghts | \data, \lambda)}{\partial \lambda} \biggr>_{\lambda}} ,
\end{equation}
where
\begin{align}
& \biggl< \frac{\partial \cost_\mathrm{bridge}(\wghts | \data, \lambda)}{\partial \lambda} \biggr>_{\lambda} = \nonumber\\
&\biggl< \sum_{i=1}^{d}{\frac{\kkone_{ii}}{2}\left(\wghtsone_i-\wghtsone_{0,i}\right)^2} - \left[\cost(\wghts | \data)  - \cost(\wghts_0 | \data)\right]   \biggr>_{\lambda} . \label{eq:ti_average}
\end{align}
Averaging in~\eqref{eq:ti_average} occurs over the distribution $\prob(\wghts)\propto e^{-\cost_\mathrm{bridge}(\wghts | \data, \lambda)} $.

In this way, TI evaluates the evidence for a model, 
\begin{equation}
\prob\left(  \mathbf{\lbl}  | \mathbf{\dpnts}, \model \right) = e^{-F}
\end{equation}
 as a correction to $\ev_0$.

\subsubsection{Experimental Setup \label{sec:ti_imp_det}}

We reuse our HMC implementation for these calculations.
HMC trajectories are all performed using $\cost_\mathrm{bridge}(\wghts | \data, \lambda)$.
Before fitting $\kk$, $dt$ is adjusted to obtain an acceptance rate in the range (0.6--0.7).
For fitting $\kk$ we set $\lambda = 0$ and perform 1000 trajectories of burn-in each $L=100$ steps long, then a further 1000 trajectories of the same length to obtain an estimate of $\kk$ according to~\eqref{eq:fit_k}.

We use 100 bridging distributions (102 distributions in total).
These distributions are sampled sequentially, from $\lambda=0$ to $\lambda=1$.
To evaluate each expectation~\eqref{eq:ti_average} 100 trajectories of HMC with $L=100$ are performed as burn-in, before a further 100 trajectories of the same length to collect samples.
We update $dt$ after every 10 bridging distributions to recover an acceptance rate in the range (0.6--0.7).
Additional trajectories, not included in any averaging or burn-in, are performed in order to set $dt$.
The integration in~\eqref{eq:ti_integrate} is performed using Simpson's rule.

\subsection{RESULTS: BAYESIAN SELECTION BETWEEN $\model^{(1)}$ AND $\model^{(3)}$  \label{sec:ti_res}}

In this section we perform a Bayesian model selection between the neural networks $\model^{(3)}$ which has three hidden layers, and $\model^{(1)}$ which has one.
These networks and their priors are fully described in Secs.~\ref{sec:data_NN} and~\ref{sec:ti_nn_and_prior}.

All calculations are performed with data set $\data_{500}$ (see Sec.~\ref{sec:data_NN}). 
To obtain $\wghts_0$  for each model we follow the standard training procedure described in Sec.~\ref{sec:rehmc}, achieving $E(\data_{500}|\wghts)=0$.

For the uniform prior~\eqref{eq:uni_prior} 
\begin{equation}
\prob\left(  \mathbf{\lbl}  | \mathbf{\dpnts}, \model \right) = \int{d\wghts e^{-E(\data|\wghts)} } \times \frac{1}{ \prod_{i=1}^d{\sigma_{i}} }  .
\end{equation}

Using TI we obtain
\begin{align*}
\log{\left( \int_{|\wghts|<\frac{\mathbf{\sigma}}{2}}{d\wghts e^{-E^{(3)}(\wghts|\data)} }   \right)} =& \left( 26.48 \pm 0.17 \right) \times 10^3   \\
\log{\left( \int_{|\wghts|<\frac{\mathbf{\sigma}}{2}}{d\wghts e^{-E^{(1)}(\wghts|\data)} }   \right)} =& \left( 19.79 \pm 0.01 \right) \times 10^3 
\end{align*}
The uncertainties shown correspond to the standard deviations of 28 independent calculations for each model.
Additionally, for our prior we have 
\begin{align*}
\log{\left(  \prod_{i=1}^d{\sigma^{(3)}_{i}}  \right)} =& 28.960 \times 10^3 \\
\log{\left(  \prod_{i=1}^d{\sigma^{(1)}_{i}}  \right)} =& 19.946 \times 10^3
\end{align*}

Thus we obtain
\begin{align}
\frac{\prob\left( \model^{(3)} |  \mathbf{\dpnts}, \mathbf{\lbl}   \right)}{\prob\left( \model^{(1)} |  \mathbf{\dpnts}, \mathbf{\lbl}   \right)} =& \frac{e^{\left( 26.475 \pm 0.173  \right) \times 10^3}}{e^{\left( 19.793 \pm 0.013  \right) \times 10^3 }} \times \frac{e^{19946}}{e^{28960}} \\
=& e^{6682} \times e^{-9014} \\
\simeq & e^{-2332} \nonumber
\end{align}

Bayes theorem clearly favours the shallow network $\model^{(1)}$ over the deep network $\model^{(3)}$.
For larger data sets or for a different prior, this preference may well be reversed.
However, for our prior and $\data_{500}$ the accuracy of $\model^{(3)}$ is not enough to justify its larger parameter space.

\section{CONCLUSION}

We have seen that for a number of small MNIST training sets sampling from the posterior does indeed lead to markedly improved generalisation than standard optimisation of the cost function.
Even better generalisation can often be obtained by sampling the temperature-adjusted posterior~\eqref{eq:thermal_posterior} at $T \neq 1$.
In particular, typical neural networks exhibit a clear minimum of the test error at $T>0$.
This suggests a stopping criterion for full batch simulated annealing of classifiers: cool until the average validation error starts to increase, then revert to the parameters with the lowest validation error.
We saw that as $T$ is increased neural network based classifiers exhibit a transition between accurate and inaccurate classification of the training data, although no sharp ``phase transition'' occurs between these states.
The absence of such a sharp transition makes sampling the temperature-adjusted posterior relatively simple and inexpensive.
Finally we showed how thermodynamic integration can be used to perform model selection on deep and shallow neural networks, avoiding the need to precisely calculate or invert the Hessian of the cost function.

\printbibliography
\end{document}